\documentclass{article}

\usepackage{amsmath}
\usepackage{amssymb}
\usepackage{graphicx}
\usepackage{wrapfig}
\usepackage{caption}
\usepackage[preprint]{neurips_2026}
\usepackage[utf8]{inputenc}
\usepackage[T1]{fontenc}
\usepackage{booktabs}
\usepackage{amsfonts}
\usepackage{nicefrac}
\usepackage{microtype}
\usepackage{xcolor}
\setcitestyle{numbers,square}

\definecolor{plum}  {rgb}{.5,0,.5}
\definecolor{forest}  {rgb}{0,.4,0} 
\definecolor{midnight}  {rgb}{0,0,.5} 
\definecolor{gray}  {rgb}{.8,.8,.8} 
\definecolor{orange}{rgb}{.87,.414,.062}
\definecolor{dkorange}{HTML}{D24E23}
\definecolor{violet}{HTML}{674ea7}
\definecolor{brick}{rgb}{.7,0,0}
\definecolor{green}{HTML}{417e27}  
\definecolor{blue}{rgb}{0,0,.7}
\definecolor{maroon}{rgb}{0.52,0,0}
\definecolor{royalpurple}{rgb}{0.47,0.32,0.66}
\definecolor{tan}{rgb}{0.82,0.70,0.54}
\usepackage[pdftex,plainpages=false,pdfpagelabels,colorlinks=true,urlcolor=midnight,linkcolor=black!80,citecolor=blue]{hyperref}

\title{CrossFlow: One-Step Generation Across Latent and Pixel Spaces}

\author{
Xiyuan Wang\textsuperscript{1},
Xiao Zhang\textsuperscript{2},
Yang Li\textsuperscript{2},
Ruoxi Jiang\textsuperscript{3},
Zhao Zhong\textsuperscript{2},
Liefeng Bo\textsuperscript{2},
Muhan Zhang\textsuperscript{1}\thanks{Corresponding author.}
\\
\textsuperscript{1}Institute for Artificial Intelligence, Peking University
\qquad
\textsuperscript{2}Tencent
\qquad
\textsuperscript{3}Fudan University
\\
\texttt{wangxiyuan@pku.edu.cn}
\qquad
\texttt{muhan@pku.edu.cn}
}

\begin{document}
\maketitle

\begin{abstract}
Most diffusion and flow-matching generators define the prior, probability path, and prediction target in the same representation space. Latent diffusion improves efficiency by moving this path into an autoencoder latent space, but the final sample is still produced by a separately trained decoder. This separation creates a mismatch: the generator is optimized for latent-space prediction, while final quality depends on how the decoder handles generated latents that may differ from clean encoder outputs. We introduce CrossFlow, a cross-space flow formulation that maps noisy latent inputs directly to pixel-space images. The key technical step is a velocity-free one-step objective: the latent trajectory defines the training path, but the supervised prediction is an image rather than a latent displacement. This lets one model act both as a one-step latent-to-pixel generator and as a decoder replacement for latent diffusion pipelines. On class-conditional ImageNet-1k at $256\times256$, CrossFlow-XL achieves 1.62 FID with one function evaluation. Ablations show that the latent encoder and pixel-space perceptual and adversarial losses are important for fidelity. These results indicate that cross-space flow objectives can combine the efficiency of latent representations with direct pixel-space supervision, without requiring a separate decoder at inference.
\end{abstract}

\section{Introduction}

Latent diffusion has become a dominant framework for image generation. Diffusion models~\citep{DDPM,NCSM} and flow-matching methods~\citep{rf,FlowMatching} provide training objectives, and Transformer backbones~\citep{dit,sit} have made these objectives scalable. Large latent diffusion systems such as Stable Diffusion 3~\citep{sd3}, FLUX~\citep{flux}, and SANA~\citep{sana} reduce the computation by running the generative trajectory in the compressed latent space of an autoencoder. This design has become standard because it avoids the cost of denoising directly in high-dimensional pixel space.

The same design also leaves an unresolved interface problem. In a latent diffusion model, the generator is trained to predict latents, while the decoder is trained to reconstruct images from latents produced by an encoder. However, at inference time, the decoder receives generated latents rather than clean encoder outputs. If these generated latents are distribution-shifted, the decoder can amplify errors. The generator also cannot be trained directly with image-level perceptual or adversarial losses unless its output is first passed through the separate decoder. Thus, latent diffusion gains efficiency by decoupling generation from decoding, but this decoupling weakens direct pixel-level supervision.

\begin{figure}[t]
    \centering
    \includegraphics[width=0.9\linewidth]{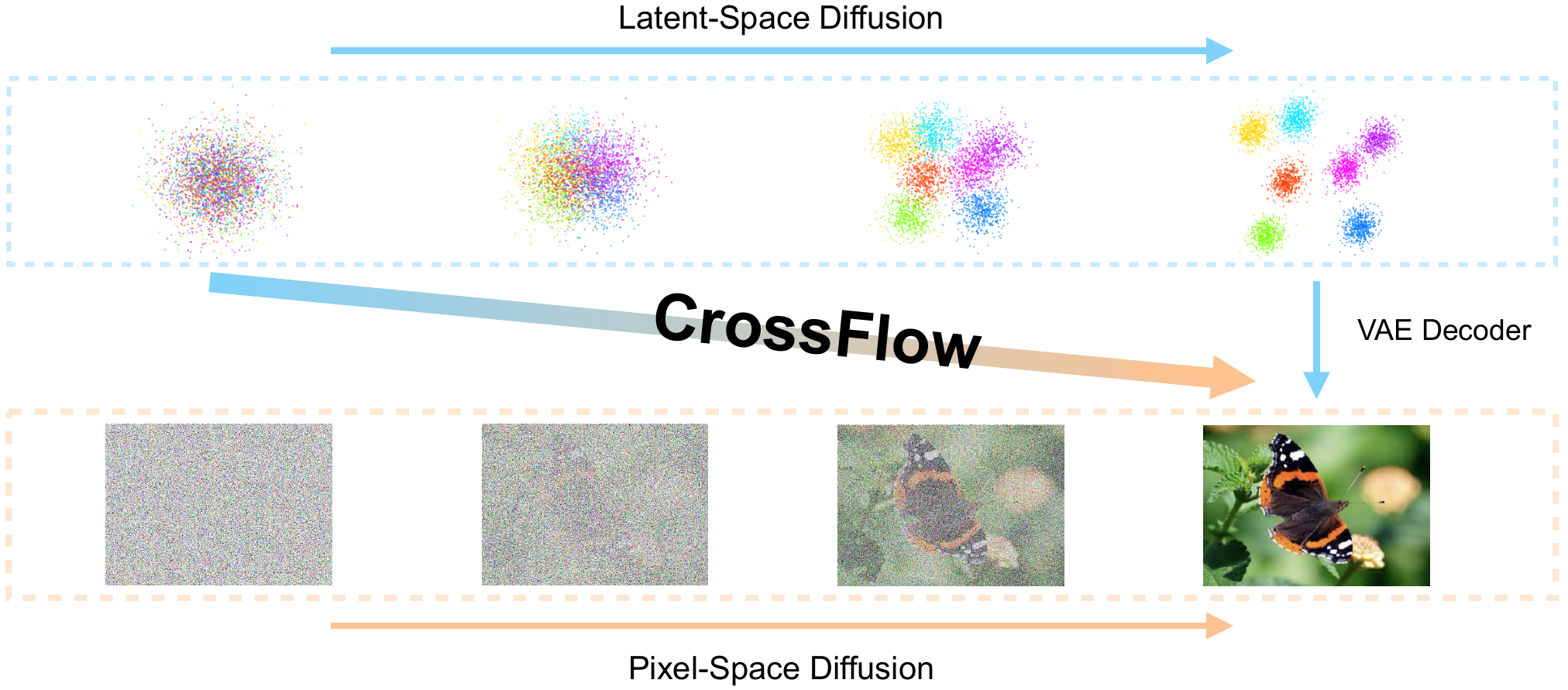}
    \caption{\textbf{CrossFlow generation paradigm.} Latent diffusion uses a two-stage pipeline: iterative denoising in latent space followed by VAE decoding. Pixel-space diffusion performs iterative denoising directly in the image domain. CrossFlow uses a single model to map a noisy latent prior directly to a pixel-space image, unifying one-step generation and latent-to-pixel decoding.}
    \label{fig:crossflow_paradigm}
\end{figure}

This paper asks a simple question: must the noise space and the output space be the same? A useful image generator could start from a latent prior, preserve the computational advantages of latent representations, and still predict the image itself. Such a model would remove the explicit decoder stage, train directly on generated images, and expose the model during training to the kind of noisy or imperfect latents that appear during generation.

We propose \emph{CrossFlow}, a cross-space one-step generation framework. As shown in Figure~\ref{fig:crossflow_paradigm}, CrossFlow maps a noisy latent input directly to a pixel-space image in a single function evaluation. This setting is different from standard flow matching, where the model predicts a velocity to update the current state, both of which must live in the same space. A latent-to-pixel model cannot update a latent variable with a pixel-space velocity. CrossFlow resolves this mismatch by deriving objectives in which the latent-space velocity is eliminated from the supervised target. The latent path still defines how clean latents are corrupted, but the network output is pixel.

% The derivation leads to two practical requirements. First, the loss cannot require the model to predict a latent-space velocity. Second, the model output must represent the image itself, rather than a quantity that is linearly combined with the latent input. These requirements yield a family of velocity-free cross-space objectives. We focus on a reconstruction-compatible branch that anchors the prediction to the clean image and propagates that image-level constraint along the noisy latent trajectory through a time-derivative term. A consistency-like branch is possible, but our diagnostic and ablation results show that the reconstruction-compatible branch is substantially more reliable.

We instantiate CrossFlow with a fixed pretrained image encoder and a Vision Transformer generator. During training, an image is encoded into a latent, corrupted along a latent trajectory, and passed to the generator, which directly predicts pixels. Because the output is already an image, we can use reconstruction, perceptual, and adversarial losses in the same training objective. The resulting model has two uses: it is a one-step latent-to-pixel generator, and it can replace the decoder in a latent diffusion pipeline whose latents are produced by a generator rather than by a clean encoder. On ImageNet-1k at $256\times256$, CrossFlow-B, CrossFlow-L, and CrossFlow-XL achieve 2.54, 1.69, and 1.62 FID, respectively, with one function evaluation. CrossFlow-XL is competitive with strong multi-step latent diffusion baselines while operating in a different regime: it maps latent noise directly to pixels and does not use a separate decoder at inference. Ablations show that both the latent representation and pixel-space auxiliary losses matter. CrossFlow also improves FID when used as a decoder for latents produced by a DiT generator in the VA-VAE latent space.

Our contributions are:
\begin{itemize}
    \item We introduce CrossFlow, a one-step generation framework in which the input prior and output image may live in different representation spaces.
    \item We implement CrossFlow as a ViT-based latent-to-pixel generator with direct pixel-space perceptual and adversarial refinement.
    \item We show strong one-step ImageNet $256\times256$ generation results and demonstrate that the same model can replace a VAE decoder for generated latents.
\end{itemize}

\section{Related Work}

CrossFlow is related to diffusion and flow-based generation, latent diffusion and autoencoding, and generative models that allow the input and output spaces to differ. The distinction from most prior work is architectural as well as objective-level: CrossFlow keeps the generative input in a compact latent space but supervises the model directly in pixel space.

\subsection{Diffusion, Flow Matching, and Few-Step Generation}

Diffusion models learn to reverse a noise-corruption process. DDPMs~\citep{DDPM} and score-based models~\citep{NCSM} established this paradigm, while DDIM~\citep{DDIM} introduced deterministic sampling for faster inference. Flow matching~\citep{FlowMatching,rf} instead learns a velocity field that transports noise to data along a prescribed path. Transformer architectures such as DiT~\citep{dit} and SiT~\citep{sit} improved scalability and have become common backbones for high-quality image generation. Many methods reduce the number of sampling steps. Consistency models~\citep{CM} learn mappings that remain consistent along probability-flow ODE trajectories, enabling one-step or few-step generation, and improved consistency training~\citep{ICT} stabilizes this objective. Progressive distillation~\citep{ProgressiveDistillation}, distribution matching distillation~\citep{DMD}, MeanFlow~\citep{meanflow}, and improved MeanFlow~\citep{imf} also compress or reformulate iterative sampling into fewer evaluations. CrossFlow shares the goal of efficient generation, but changes the supervised space. These methods typically define the trajectory, velocity, and prediction target in the smae space; CrossFlow takes a noisy latent-space input while predicting a pixel-space image.

\subsection{Latent Diffusion and Autoencoders}

Latent diffusion models~\citep{ldm} reduce generation cost by running diffusion in the latent space of a pretrained autoencoder rather than directly in pixel space~\citep{adm}. This strategy underlies many modern systems, including SDXL~\citep{sdxl}, Stable Diffusion 3~\citep{sd3}, FLUX~\citep{flux}, and SANA~\citep{sana}. The autoencoder is therefore a core part of the system, and methods such as VQ-VAE~\citep{vqvae}, VQ-GAN~\citep{vqgan}, and DC-AE~\citep{dcae} study the trade-off among compression, reconstruction fidelity, and efficiency. A related line of work improves compatibility between autoencoders and latent generators. LSGM~\citep{LSGM} proposed score-based generative modeling in latent space with joint optimization; REPA-E~\citep{REPA-E} tuned VAEs with latent diffusion Transformers through representation alignment; DSD~\citep{DSD} interpreted diffusion as self-distillation for end-to-end latent diffusion; and unified-latent VAEs~\citep{ULVAE,UNITEVAE} co-optimize the tokenizer and diffusion model. These methods address the mismatch between tokenization and generation, but they generally retain separate encoder, generator, and decoder modules. CrossFlow explores a complementary decomposition: the encoder may remain fixed, while the generator and decoder are merged into one model trained on noisy latent inputs and pixel-space targets.

\subsection{Generation Across Different Spaces}

Natural images have strong spatial and semantic structure, so their intrinsic dimension can be lower than their ambient pixel dimension. GANs map low-dimensional noise directly to images~\citep{GAN,biggan,stylegan}, but standard normalizing flows~\citep{NormalizingFlows,RealNVP} require invertible transformations and therefore impose dimension matching. Augmented normalizing flows~\citep{AugmentedNF} and injective flows~\citep{InjectiveFlows} relax this restriction in different ways. Standard diffusion and flow-matching models~\citep{DDIM,DDPM,NCSM,FlowMatching} usually avoid the issue by defining the full trajectory inside a single space. CrossFlow targets the latent-to-pixel case. It does not require an invertible map and does not decode a generated latent through a separately trained module. Instead, the latent path is used only to define noisy inputs and time derivatives, while the supervised prediction is an image. % Cross-modal generation is another setting where input and output spaces differ~\citep{multimodalflow1,multimodalflow2}; many such systems introduce a shared representation between modalities. CrossFlow addresses a narrower but practically important question: how to obtain a flow-style training objective when the noisy input remains latent-valued and the desired output is pixel-valued.

\section{Method}
Diffusion models are typically confined to a single operating space. As a consequence, they lack native support for cross-space generation, such as directly mapping from noisy latents to pixels. This restriction is particularly pronounced in latent diffusion models (LDMs), which necessitate an auxiliary decoder to project the generated latents back into the pixel domain. We propose \textit{CrossFlow} to bridge this gap by extending flow-based supervision to cross-space configurations. By optimizing the model to map a latent prior directly to a sample in the image space, CrossFlow eliminates the need for sequential latent generation and separate decoding.

\subsection{Background: Latent Diffusion Models (LDM)}

Consider a data sample $x \sim p_{\text{data}}$ and Gaussian noise $\epsilon \sim \mathcal{N}(0, I)$. The LDM first maps the data into a latent representation via an encoder $z = \mathcal{E}(x)$, and subsequently constructs a noisy latent trajectory:
\begin{equation}
z_t = \alpha(t) z + \beta(t) \epsilon,
\label{eq:ldm_forward}
\end{equation}
where $t \in [0, 1]$ parameterizes the noise schedule. The coefficient functions $\alpha(t)$ and $\beta(t)$ satisfy the boundary conditions $\alpha(0) = 1$, $\beta(0) = 0$, $\alpha(1) = 0$, and $\beta(1) = 1$. Generative modeling in an LDM proceeds by solving the probability-flow ordinary differential equation (PF-ODE):
\begin{equation}
z_0 = z_1 + \int_{1}^{0} v(z_\tau, \tau)\,\mathrm{d}\tau,
\label{eq:pf_ode}
\end{equation}
where $v(z_{\tau}, \tau)$ denotes the marginal velocity of the latent state $z_{\tau}$. The resulting clean latent $z_0$ is subsequently mapped back to the pixel space using a decoder, yielding $x_0 = \mathcal{D}(z_0)$. To estimate the intractable marginal velocity $v(z_t, t)$, we leverage the tractable conditional velocity $v(z_t, t, z)$:
\begin{equation}
v(z_t, t, z) = \dot{\alpha}(t)z + \dot{\beta}(t)\epsilon = \left(\dot{\alpha}(t)-\frac{\dot{\beta}(t)}{\beta(t)}\alpha(t)\right)z + \frac{\dot{\beta}(t)}{\beta(t)}z_t,
\label{eq:conditional_velocity}
\end{equation}
where dots indicate time derivatives with respect to $t$. The diffusion model $F_\theta$ is optimized via score-matching regression:
\begin{equation}
\mathcal{L}_{v} = \mathbb{E}_{x,\epsilon,t} \left\| F_\theta(z_t, t) - v(z_t, t, z) \right\|_2^2.
\end{equation}
At convergence, $F_{\theta}(z_t,t)$ recovers the marginal velocity $v(z_t,t)$ via the conditional expectation:
\begin{equation}
v(z_t, t) = \mathbb{E}_{z\sim p(\cdot|z_t)}\left[v(z_t, t, z)\right].
\label{eq:marginal_velocity}
\end{equation}

During inference, Eq.~\eqref{eq:pf_ode} is integrated numerically to transport $z_t$ toward the clean latent $z_0$. This formulation strictly mandates that the model output $F_\theta(z_t, t)$ resides in the same mathematical space as $z_t$, fundamentally preventing standard LDMs from executing end-to-end latent-to-pixel generation.

One-step generative models relax the integration overhead by learning a direct mapping from the latent prior $z_1$ to the clean target $z_0$ within a single evaluation. However, because their underlying derivations remain tied to intra-space velocity fields, they cannot be readily extended to cross-space generation. To overcome this limitation, we introduce a unified framework for cross-space flow supervision.

\subsection{A Unified Framework for Cross-Space Flow Supervision}

In principle, one-step models map between arbitrary points along a probability path without explicit numerical integration. To analyze these methodologies systematically, we first introduce a unified formulation that encompasses a broad family of single-step generative paradigms. For $0 \le r \le t \le 1$ and arbitrary coefficient functions $\lambda(t, r)$ and $\gamma(t, r)$, we define the target function:
\begin{equation}
\mathcal{Z}(z_t, t, r; \lambda, \gamma) = \lambda(t, r) z_t + \frac{1}{\gamma(t, r)} \int_{t}^{r} v(z_\tau, \tau)\,\mathrm{d}\tau.
\label{eq:general_z}
\end{equation}
This formulation highlights how varying the parameterization of the ODE flow recovers different architectures. In the first regime, where $\lambda(t, r)=\gamma(t, r)=1$, $\mathcal{Z}$ corresponds to the exact solution of the PF-ODE. Setting $r=0$ recovers consistency models~\citep{CM,ICT}, whereas specifying a flexible destination $r \in [0, 1]$ yields consistency trajectory models~\citep{CTM}. This ODE-based parameterization is also closely linked to shortcut models~\citep{shortcut}. In a second regime, where $\lambda=0$, $\mathcal{Z}$ simplifies to a time-averaged velocity integral. Specifically, choosing $\gamma(t, r)=r-t$, $\alpha(t)=1-t$, and $\beta(t)=t$ instantiates MeanFlow~\citep{meanflow} and Improved MeanFlow~\citep{imf}.

Despite its generality, this unified representation reveals why existing methods are incompatible with cross-space generation: their training objectives depend inherently on the latent-space velocity field. For instance, few-step models like Improved MeanFlow~\citep{imf} supervise the network by matching a linear combination of the model output and its time derivative to a local latent-velocity target. Such formulations rely entirely on the structural homogeneity between the input and output spaces. To lift this constraint, we introduce two design criteria.

\paragraph{Criterion 1: Eliminating Latent-Velocity Dependencies.}
To enable cross-space supervision, the objective function must be structurally decoupled from latent-space velocities. Direct supervision via velocity terms prevents parameterizing a generator that maps latents to pixels. 

Following standard conventions~\citep{meanflow}, we differentiate Eq.~\eqref{eq:general_z} with respect to $t$. Simplifying the expression via Eq.~\eqref{eq:conditional_velocity} and Eq.~\eqref{eq:marginal_velocity} yields:
\begin{equation}
\begin{aligned}
\partial_t\gamma(t,r)\mathcal{Z}(z_t, t, r; \lambda, \gamma) + \gamma(t,r)\frac{\mathrm{d}\mathcal{Z}}{\mathrm{d}t} 
&= \partial_t (\gamma(t, r) \lambda(t, r))\left(\alpha(t)-\dot{\alpha}(t)\frac{\beta(t)}{\dot{\beta}(t)}\right)\mathbb{E}_{z_0\sim p(\cdot | z_t)}[z_0] \\
&+\left(\partial_t (\gamma(t, r) \lambda(t, r))\frac{\beta(t)}{\dot{\beta}(t)}+\gamma(t, r) \lambda(t, r)-1\right)v(z_t,t).
\end{aligned}
\label{eq:z_derivative}
\end{equation}
To cancel the problematic marginal velocity term $v(z_t, t)$ on the right-hand side, we enforce the following structural constraint:
\begin{equation}
\partial_t(\gamma(t, r) \lambda(t, r))\frac{\beta(t)}{\dot{\beta}(t)}+\gamma(t, r) \lambda(t, r)-1 = 0 \implies \gamma(t,r)\lambda(t, r)=1-\frac{\phi(r)}{\beta(t)},
\label{eq:cancel_v}
\end{equation}
where $\phi(r)$ is a function dependent solely on the destination timestep $r$. This provides our first necessary condition for cross-space supervision.

\paragraph{Criterion 2: Endpoint Boundary Conditions at $r=0$.}
We additionally require the framework to yield a valid image at the destination boundary $r = 0$, which mirrors the endpoint of the underlying PF-ODE:
\begin{equation}
\mathcal{Z}(z_t, t, 0; \lambda, \gamma) = z_t + \int_{t}^{0} v(z_\tau, \tau)\,\mathrm{d}\tau.
\label{eq:sample_target}
\end{equation}
Aligning Eq.~\eqref{eq:sample_target} with the general formulation in Eq.~\eqref{eq:general_z} mandates the boundary conditions $\lambda(t, 0)=1$ and $\gamma(t, 0)=1$.

\paragraph{Training Objective.}
Synthesizing the generalized trajectory target in Eq.~\eqref{eq:z_derivative} with our design criteria yields a flexible family of cross-space optimization objectives. We replace the latent-space representation $\mathcal{Z}$ in Eq.~\eqref{eq:z_derivative} with a pixel-space neural network $F_\theta(z_t,t,r)$ and optimize it to reconstruct the paired pixel-space target $x_0$. This yields the general cross-space loss function:
\begin{equation}
\small
\mathcal{L}_{\text{General}} = \mathbb{E}_{x_0,\epsilon,t,r} \left\| \partial_t \gamma(t,r)\, F_\theta(z_t,t,r) + \gamma(t,r)\frac{\mathrm{d} F_\theta(z_t,t,r)}{\mathrm{d}t} - \frac{\phi(r)\dot{\beta}(t)}{\beta(t)^2} \left( \alpha(t)-\dot{\alpha}(t)\frac{\beta(t)}{\dot{\beta}(t)} \right)x_0 \right\|_2^2,
\label{eq:crossflow_loss_family}
\end{equation}
where $\frac{\mathrm{d} F_\theta}{\mathrm{d}t}$ denotes the total derivative evaluated along the latent diffusion trajectory. This objective is governed by four schedule functions. The pair $\alpha(t)$ and $\beta(t)$ determines the latent diffusion path, while $\gamma(t,r)$ and $\phi(r)$ isolate the specific cross-space statistics learned by the network. To satisfy Criterion 2, we enforce $\gamma(t,0)=1$ and $\phi(0)=0$. Together, these boundary conditions guarantee that evaluating $F_\theta(\epsilon, 1, 0)$ with a standard normal prior $\epsilon \sim \mathcal{N}(0, I)$ maps directly to a valid pixel-space sample.

For our primary empirical evaluations, we adopt a simple and intuitive default instantiation: a linear noise schedule given by $\alpha(t)=1-t$ and $\beta(t)=t$, alongside $\phi(r)=r$ (satisfying $\phi(0)=0$) and $\lambda(t,r)=1$. Substituting these choices into Eq.~\eqref{eq:cancel_v} uniquely determines $\gamma(t,r)=1-\frac{r}{t}$. Plugging these schedules into Eq.~\eqref{eq:crossflow_loss_family} yields the concrete CrossFlow objective utilized in our experiments:

\begin{equation}
\mathcal{L}_{\mathrm{CF}} = \mathbb{E}_{x_0,\epsilon,t,r} \left\| \frac{r}{t^2} \cdot \left( F_\theta(z_t,t,r) + \frac{t(t-r)}{r} \frac{\mathrm{d} F_\theta(z_t,t,r)}{\mathrm{d}t} \, - \, x_0 \right) \right\|_2^2.
\label{eq:crossflow_loss}
\end{equation}

Crucially, $F_\theta$ is never optimized to predict a latent-space displacement vector. Instead, it natively outputs a clean image while its trajectory dynamics along the latent space are structurally bound by the cross-space flow loss. We explore alternative valid configurations of $\alpha, \beta, \gamma,$ and $\phi$ across synthetic toy domains in Appendix~\ref{app:crossflow_design_space}.

\subsection{Implementation Details}

\textbf{Computing the time derivative.} Following Improved MeanFlow~\citep{imf}, the total time derivative $\frac{\mathrm dF_{\theta}}{\mathrm dt}$ is evaluated along the latent diffusion trajectory via a Jacobian-Vector Product ($\mathrm{JVP}$):
$\frac{\mathrm dF_{\theta}}{\mathrm dt} = \mathrm{JVP}\Big(F_\theta(z_t, t, r), , \big(v(z_t, t, z), 1, 0\big)\Big),$
where the first tuple defines the function inputs and the second tuple represents the corresponding tangent directions. To optimize computational efficiency, we execute this step using forward-mode automatic differentiation (\texttt{torch.func.jvp}), which incurs a marginal overhead of approximately twice the runtime of a standard forward pass. Following the convention established by MeanFlow~\citep{meanflow}, the $\mathrm{JVP}$ term is detached from the computational graph during backpropagation.

\textbf{Two-Stage Training Pipeline.} In addition to the generator $F_{\theta}$, our framework relies on an encoder $\mathcal{E}$ to project input images into the latent domain. Although the cross-space objective theoretically supports end-to-end joint optimization of $F_{\theta}$ and $\mathcal{E}$, simultaneous training exhibits optimization instabilities in practice. To mitigate this issue, we adopt a decoupled, two-stage training pipeline. In the first stage, the encoder $\mathcal{E}$ is trained within a standard autoencoder framework using a combination of reconstruction and adversarial objectives; the corresponding decoder is subsequently discarded. In the second stage, the weights of $\mathcal{E}$ are frozen, and the cross-space generator $F_{\theta}$ is trained from scratch. This modular formulation allows CrossFlow to seamlessly integrate high-performance, pretrained structural encoders. By default, we employ the encoder from VA-VAE~\citep{VAVAE}. We defer the development of stable single-stage joint training strategies to future work.
% # we do use a pretrained VAE in our primary exps

\textbf{Loss function.} Because the CrossFlow objective directly aligns the network outputs and their time derivatives with the pixel-space target $x_0$, the framework can natively leverage advanced structural and perceptual image metrics to guide generation. Mirroring standard practice in autoencoder training, our total optimization objective augments the pixel-wise $L1$ loss with an adversarial GAN loss and perceptual loss. Notably, incorporating the analytical weight $\frac{r}{t^2}$ into the perceptual loss induces training instability. We therefore omit this weighting factor during perceptual loss computation.

\section{Experiments}\label{sec:experiments}

\begin{figure}[!t]
\centering
\includegraphics[width=0.9\textwidth]{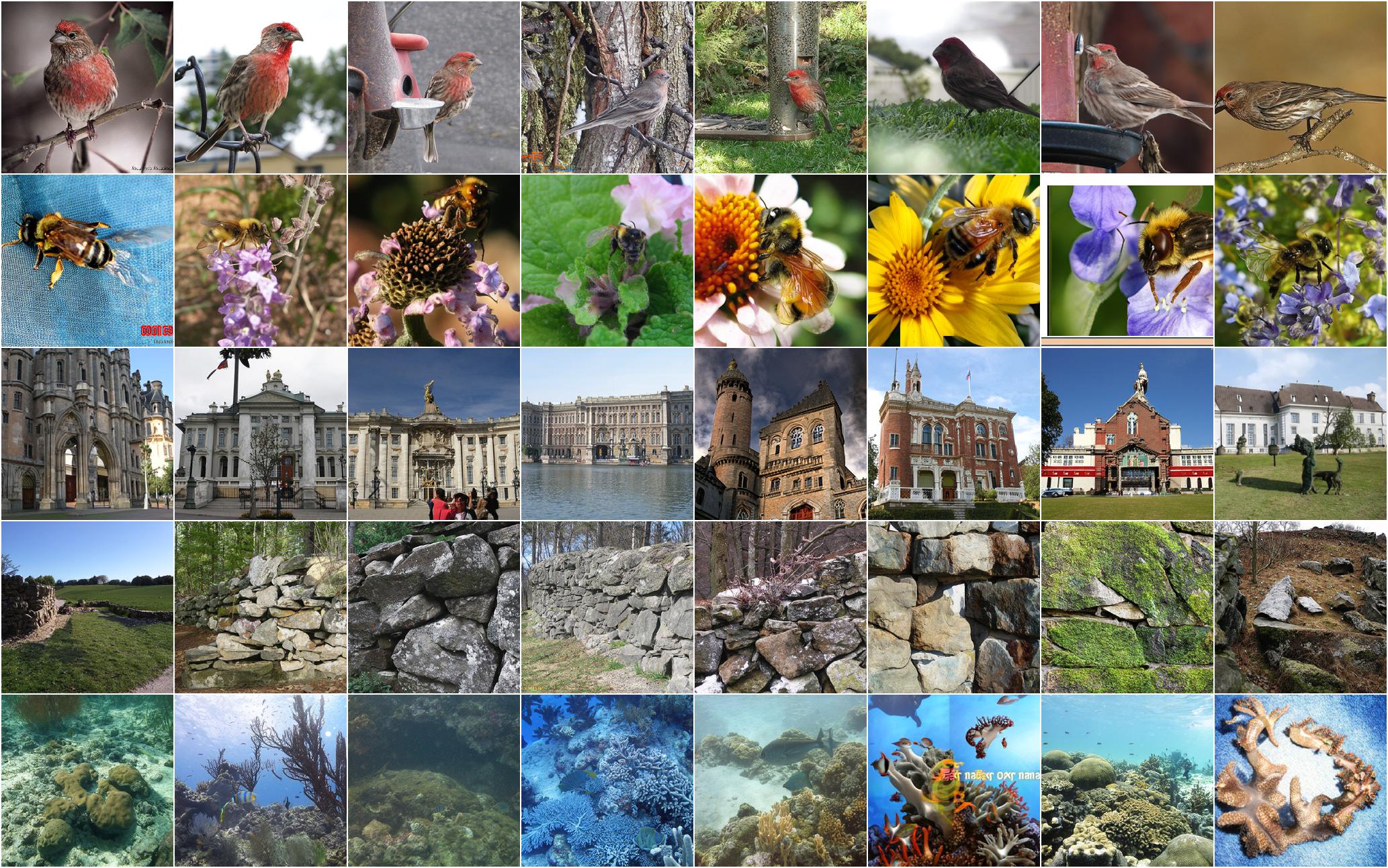}
\caption{\textbf{Uncurated 1-NFE class-conditional samples on ImageNet $256\times256$.} Samples are generated by CrossFlow-XL from latent noise directly to pixels. Rows correspond to class 12 (house finch, linnet, \emph{Carpodacus mexicanus}), class 309 (bee), class 698 (palace), and class 973 (coral reef), respectively.}
\label{fig:imagenet_samples}
\end{figure}

We evaluate CrossFlow on ImageNet-1k~\citep{imgnet} at resolution $256\times256$. We follow the standard class-conditional protocol and report Fr\'{e}chet Inception Distance (FID)~\citep{fid} on 50,000 generated samples. The main evaluation is one-function-evaluation generation (1-NFE): the model receives a sample from the latent prior and outputs pixels directly. Unless otherwise stated, we use the linear schedule $\alpha(t)=1-t$, $\beta(t)=t$.

We implement $F_{\theta}$ with a Vision Transformer (ViT) backbone~\citep{ViT}. Following ViT conventions, we evaluate three variants: \textbf{CrossFlow-B} with 12 layers and hidden dimension 768, \textbf{CrossFlow-L} with 24 layers and hidden dimension 1024, and \textbf{CrossFlow-XL} with 28 layers and hidden dimension 1152. All models are trained using the Muon optimizer~\citep{muon,KimiMuon}, with a learning rate=8e-4 and epochs=160.

\begin{table}[t]
\centering
\caption{\textbf{System-level comparison on ImageNet $256\times256$ generation.} FID is evaluated on 50,000 generated samples. $\times 2$ in NFE indicates classifier-free guidance. Methods are grouped by sampling regime and generation space; CrossFlow is a one-step latent-to-pixel generator.}
\vspace{0.5em}
\scriptsize
\setlength{\tabcolsep}{4.5pt}
\begin{tabular}{l | cc | c | c| l | cc |c |c}
\toprule
\textbf{Method} & NFE & Space & Params & FID $\downarrow$ & \textbf{Method} & NFE & Space & Params & FID $\downarrow$\\
\midrule
\multicolumn{5}{l|}{\textit{Multi-step latent-space diffusion/flow}} & \multicolumn{5}{l}{\textit{1-NFE pixel-space GAN}}\\
\quad DiT-XL/2~\citep{dit} & 250$\times$2 & latent & 675M & 2.27 & \quad BigGAN-deep~\citep{biggan} & 1 & pixel & 56M & 6.95\\
\quad SiT-XL/2~\citep{sit} & 250$\times$2 & latent & 675M & 2.06 & \quad StyleGAN-XL~\citep{stylegan} & 1 & pixel & 166M & 2.30 \\
\quad SiT-XL/2 + REPA~\citep{repa} & 250$\times$2 & latent & 675M & \textbf{1.42} & \quad GigaGAN~\citep{gigagan} & 1 & pixel & 569M & 3.45\\
\midrule
\multicolumn{5}{l|}{\textit{Multi-step pixel-space diffusion/flow}} & \multicolumn{5}{l}{\textit{1-NFE pixel-space diffusion/flow}}\\
\quad ADM-G~\citep{adm} & 250$\times$2 & pixel & 554M & 4.59 & \quad EPG-L/16~\citep{epg} & 1 & pixel & 540M & 8.82 \\
\quad SiD, UViT~\citep{sid} & 1000$\times$2 & pixel & 2.5B & 2.44 & \quad pMF-B/16~\citep{pmf} & 1 & pixel & 118M & 3.12\\
\quad VDM++~\citep{vdmpp} & 256$\times$2 & pixel & 2.5B & 2.12 & \quad pMF-L/16~\citep{pmf} & 1 & pixel & 410M & 2.52\\
\quad SiD2~\citep{sid2} & 512$\times$2 & pixel & -- & 1.38 & \quad pMF-H/16~\citep{pmf} & 1 & pixel & 956M & 2.22 \\
\quad PixelDiT-XL/16~\citep{pixeldit} & 100$\times$2 & pixel & 797M & \textbf{1.61} \\
\midrule
\multicolumn{5}{l|}{\textit{1-NFE latent-space diffusion/flow}} & \multicolumn{5}{l}{\textbf{\textit{1-NFE latent-to-pixel (ours)}}} \\
\quad iCT-XL/2~\citep{ICT} & 1 & latent & 675M  & 34.24 & \quad CrossFlow-B & 1 & latent$\to$pixel & 95M & 2.54 \\
\quad Shortcut-XL/2~\citep{shortcut} & 1 & latent & 676M & 10.60 & \quad CrossFlow-L & 1 & latent$\to$pixel & 315M & 1.69 \\
\quad MeanFlow-XL/2~\citep{meanflow} & 1 & latent & 676M & 3.43 & \quad CrossFlow-XL & 1 & latent$\to$pixel & 461M & \textbf{1.62}\\
\quad iMF-XL/2~\citep{imf} & 1 & latent & 659M & 1.72 \\
\bottomrule
\end{tabular}
\label{tab:main_results}
\end{table}

\subsection{Main Results: 1-NFE FID}

We first evaluate one-step image generation. Table~\ref{tab:main_results} compares CrossFlow against prior ImageNet $256\times256$ results, across different representation spaces and sampling regimes.
%, the table is a system-level landscape comparison.
%including multi-step latent-space methods, multi-step pixel-space methods, one-step latent-space methods, GANs, one-step pixel-space diffusion or flow models, and our one-step latent-to-pixel setting.

Overall, CrossFlow achieves strong results with a single function evaluation.  CrossFlow-XL achieves 1.62 FID, a competitive performance with multi-step latent diffusion methods such as DiT-XL/2 (2.27 FID with 250 steps) and SiT-XL/2 (2.06 FID with 250 steps). Among the one-step methods, our CrossFlow improves over the one-step latent-space method iMF-XL/2 (1.72 FID), while avoiding a separate decoder at inference. We provide uncrated samples in Figure~\ref{fig:imagenet_samples} for a qualitative comparison.

We also demonstrate the scalability of the CrossFlow model: the FID improves from 2.54 (CrossFlow-B) to 1.62 (CrossFlow-XL) as we increase the parameters from 95M to 461M. 

%as the parameter number increases from 95M to 461 
%CrossFlow-B reaches 2.54 FID, CrossFlow-L reaches 1.69 FID, and CrossFlow-XL reaches 1.62 FID.

\begin{figure}[!t]
\centering
\includegraphics[trim={0 620 0 0}, clip, width=0.9\textwidth]{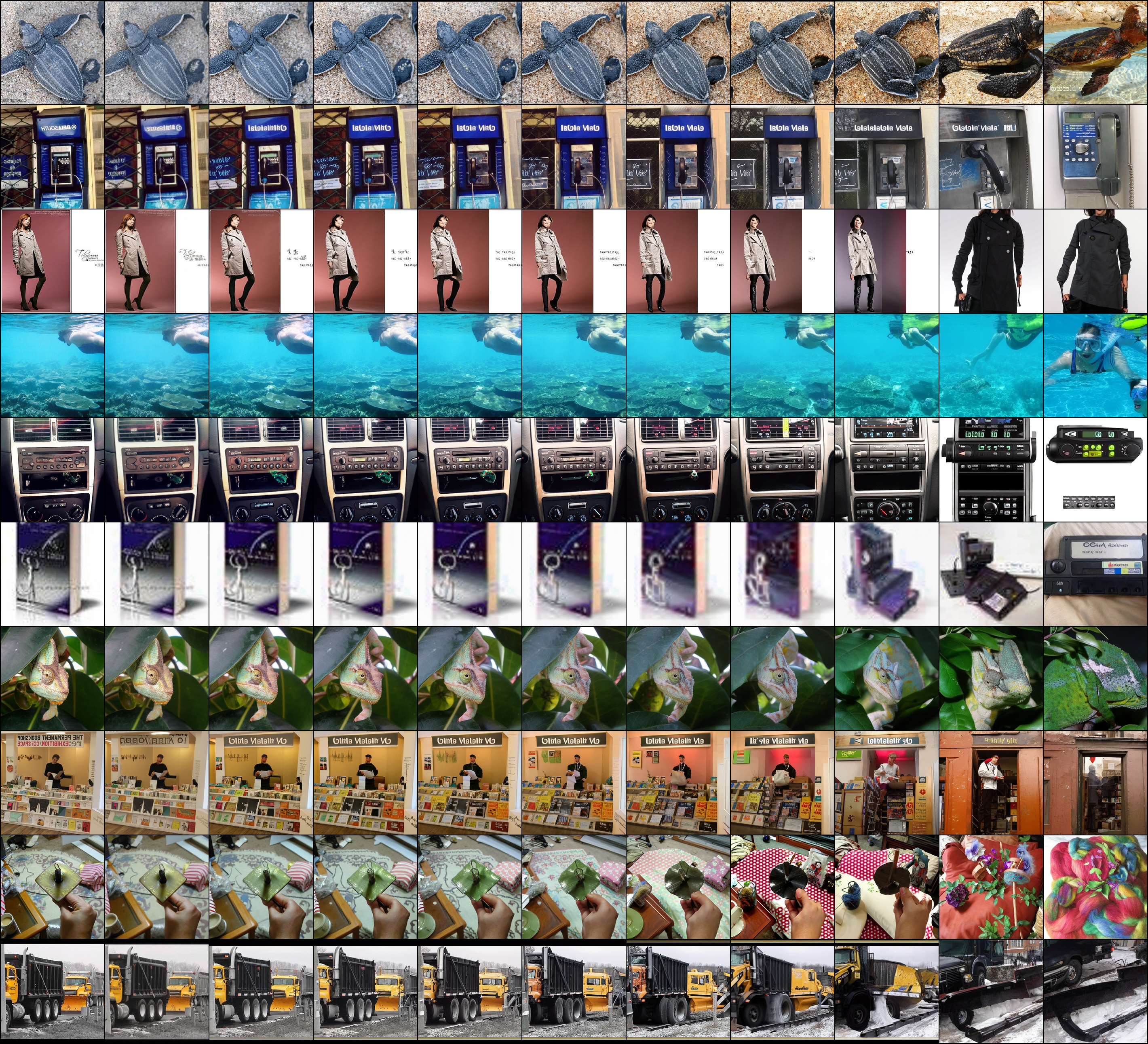}
\caption{\textbf{Visualization of $F_{\theta}(z_t, t, 0)$ as $t$ varies from 0 (left, clean latent) to 1 (right, pure noise).} Each row represents a different semantic category. The interpolation preserves recognizable structure while progressively introducing stochastic variation and noise.}
\label{fig:imagenet_interp}
\end{figure}

\begin{table}[!ht]
\centering
\caption{\textbf{Ablation studies.} Unless otherwise specified, all models use CrossFlow-B, the VA-VAE encoder, the full loss configuration, the Muon optimizer, $r$-only time conditioning, no class-label dropout, and 160 training epochs.}
\vspace{0.5em}
\small
\setlength{\tabcolsep}{3pt}
\begin{tabular}{l c l c}
\toprule
\textbf{Ablation} & \textbf{FID $\downarrow$} & \textbf{Ablation} & \textbf{FID $\downarrow$}\\
\midrule
\multicolumn{2}{l}{\textit{Reference setting}} \\
Full CrossFlow-B & \textbf{2.54} \\
\midrule
\multicolumn{2}{l}{\textit{Encoder choice}} \\
Encoder: SD-VAE~\citep{ldm}, 16 latent channels & 4.08 &
Encoder: FLUX.2~\citep{flux2}, 128 latent channels & 3.81 \\
\midrule
\multicolumn{2}{l}{\textit{Auxiliary losses and objective variants}} \\
Loss: $\phi(r) = 0$ & 8.70 &
Loss: JVP with estimated marginal velocity & 7.06\\
Loss: without GAN loss & 8.86 &
Loss: without perceptual loss & 30.21 \\
Loss: without both perceptual and GAN losses & 50.65 \\
\midrule
\multicolumn{2}{l}{\textit{Training and conditioning choices}} \\
Optimizer: Muon $\rightarrow$ AdamW & 2.80 &
Time conditioning: $r\rightarrow(r,t)$ & 2.73 \\
Class conditioning: drop class labels by 10\% & 2.76 \\
\bottomrule
\end{tabular}
\label{tab:ablation}
\end{table}

\subsection{Ablation Studies}
We ablate our primary design choices using the CrossFlow-B architecture. As shown in Table~\ref{tab:ablation}, these experiments are categorized into encoder selection, auxiliary losses and objective variants, and training or conditioning configurations. Unless otherwise specified, our baseline reference setting utilizes the VA-VAE encoder, the full CrossFlow objective (including both perceptual and GAN losses), the Muon optimizer, $r$-only time conditioning, and no class-label dropout. 

\textbf{Encoder Selection.} Our framework leverages the latent space of a pretrained VAE. In this suite, we compare our default VA-VAE~\citep{VAVAE} (32 latent channels) against SD-VAE~\citep{ldm} (16 latent channels) and FLUX.2~\citep{flux2} (128 latent channels). 
The results indicate that the VA-VAE latent space yields the strongest performance, achieving an FID of 2.54, compared to 3.81 for FLUX.2 and 4.08 for SD-VAE. This trend aligns directly with the established gFID rankings of these VAEs on ImageNet-1k. These findings suggest that the specific structure of the latent representation could contribute to direct, single-step latent-to-pixel generation—though this experiment does not entirely isolate individual contributing factors, such as latent channel capacity.

\textbf{Loss and Objective Variants.} We next evaluate the impact of individual training objectives. Removing the adversarial (GAN) loss degrades the FID from 2.54 to 8.86, while omitting the perceptual loss causes a much more severe drop to 30.21. Excluding both auxiliary losses yields a poor FID of 50.65. The cross-space flow objective provides the necessary structural training signal for single-step generation, whereas the perceptual and adversarial losses act as vital regularizers that dramatically boost high-frequency sample fidelity.  

Furthermore, setting $\phi(r) = 0$ results in an FID of 8.70, and utilizing a JVP (Jacobian-Vector Product) formulation with estimated marginal velocity yields 7.06.

Crucially, these results do not imply that the core CrossFlow loss is secondary. While perceptual and GAN losses are standard staples in image VAE training~\citep{vqgan}, a traditional VAE decoder cannot natively generate samples directly from noise. The cross-space flow objective provides the necessary structural training signal for single-step generation, whereas the perceptual and adversarial losses act as vital regularizers that dramatically boost high-frequency sample fidelity. We explore the sensitivity of the GAN loss weight further in Figure~\ref{fig:gan_weight} and provide comprehensive details in Appendix~\ref{app:GANweight}.

\textbf{Training and Conditioning Choices.} The remaining ablations demonstrate that while our default optimization choices are beneficial, they are less critical to overall performance than the core loss configuration. Swapping the Muon optimizer for AdamW increases the FID to 2.80. Transitioning from $r$-only time conditioning to full $(r, t)$ conditioning yields an FID of 2.73, while introducing a 10\% class-label dropout rate results in an FID of 2.76.

\subsection{Ablation Studies}
We ablate the main design choices using CrossFlow-B. Table~\ref{tab:ablation} groups the experiments into encoder choice, objective and auxiliary losses, and training or conditioning choices. Unless otherwise specified, the reference setting uses the VA-VAE encoder, the CrossFlow objective with perceptual and GAN losses, the Muon optimizer, $r$-only time conditioning, and no class-label dropout. This organization separates representation choices from loss choices and secondary training choices, making the relative effect of each design axis easier to read.

\textbf{Encoder ablation:}. Our model leverages the encoder from a pretrained VAE. Here, we compare the performance between our default VA-VAE~\citep{VAVAE} with SD-VAE~\citep{ldm} and FLUX.2~\citep{flux2}.
The encoder ablation shows that the VA-VAE latent space gives the best result among the tested encoders. FLUX.2 reaches 3.81 FID and SD-VAE reaches 4.08 FID, compared with 2.54 FID for VA-VAE. This is consistent with the ranking of GFID of those VAE in ImageNet-1k. 
This suggests that the latent representation matters for direct one-step latent-to-pixel generation, although the experiment does not isolate a single cause, such as channel count alone.

\textbf{Loss ablation:} 
We first ablate the choice of training loss, Removing the GAN loss degrades FID from 2.54 to 8.86, while removing the perceptual loss causes a larger degradation to 30.21. Removing both auxiliary losses gives 50.65 FID. These results should not be interpreted as saying that the CrossFlow loss is incidental. Perceptual and GAN losses are common in image VAE training~\citep{vqgan}, but a standard VAE decoder cannot generate from noise. The consistency-loss baseline also shows that the cross-space flow objective is needed for one-step generation. Overall, the CrossFlow objective supplies the structural training signal, while perceptual and adversarial losses substantially improve sample fidelity. Figure~\ref{fig:gan_weight} further shows that the GAN loss weight matters; Appendix~\ref{app:GANweight} provides details.

The remaining ablations show that the default training choices are meaningful but less dominant than the objective and loss configuration. Replacing Muon with AdamW gives 2.80 FID, using $(r,t)$ time conditioning gives 2.73 FID, and dropping class labels with 10\% probability gives 2.76 FID.

\begin{minipage}[t]{1.0\linewidth}
\begin{minipage}[t]{0.48\linewidth}
    \centering
    \includegraphics[width=1.0\linewidth]{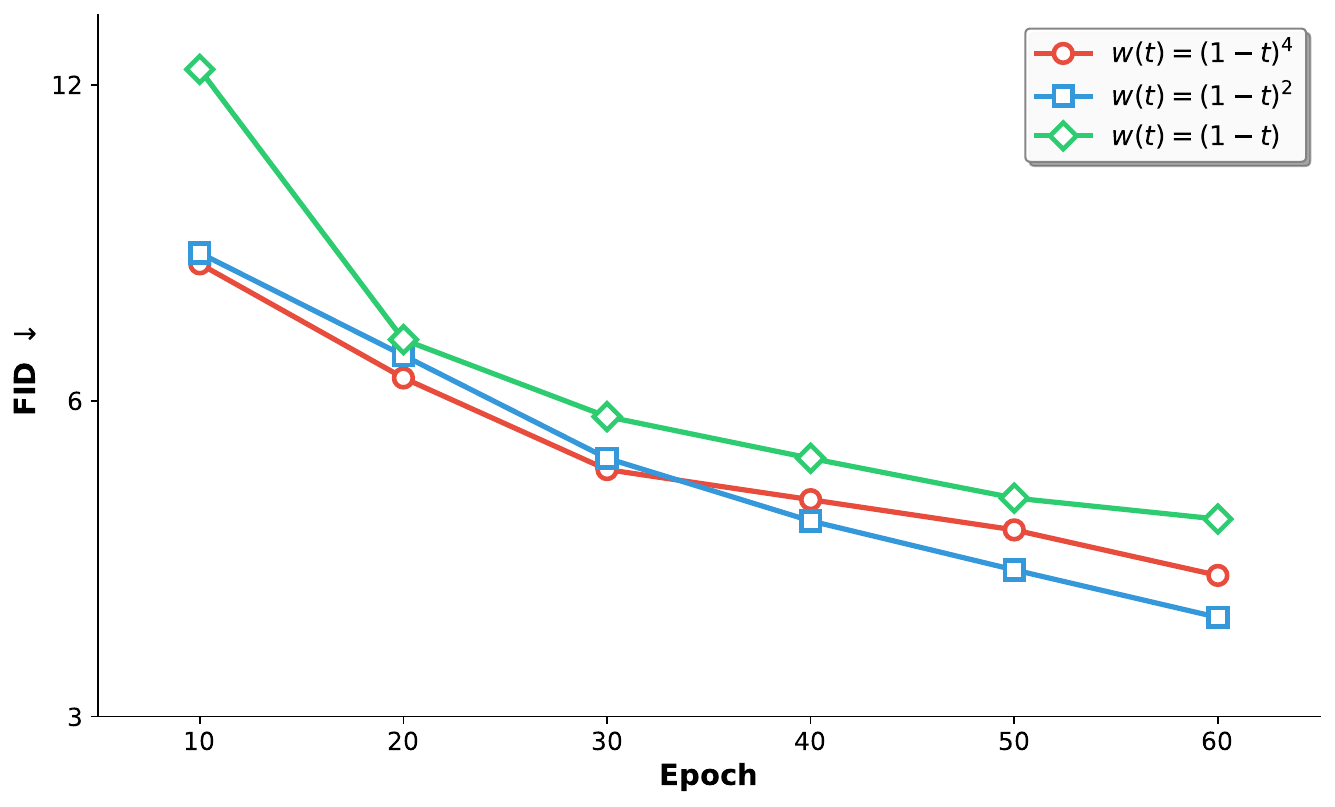}
    \captionof{figure}{\textbf{Ablation of the GAN weight scheduler $w(t)$.} Higher GAN weights lead to higher FID because of partial mode collapse, while overly small weights on high-noise latents slow initial convergence. The optimized schedule balances stability and convergence speed.}
    \label{fig:gan_weight}
\end{minipage}
\hfill
\begin{minipage}[t]{0.48\linewidth}
    \centering
    \includegraphics[width=1.0\linewidth]{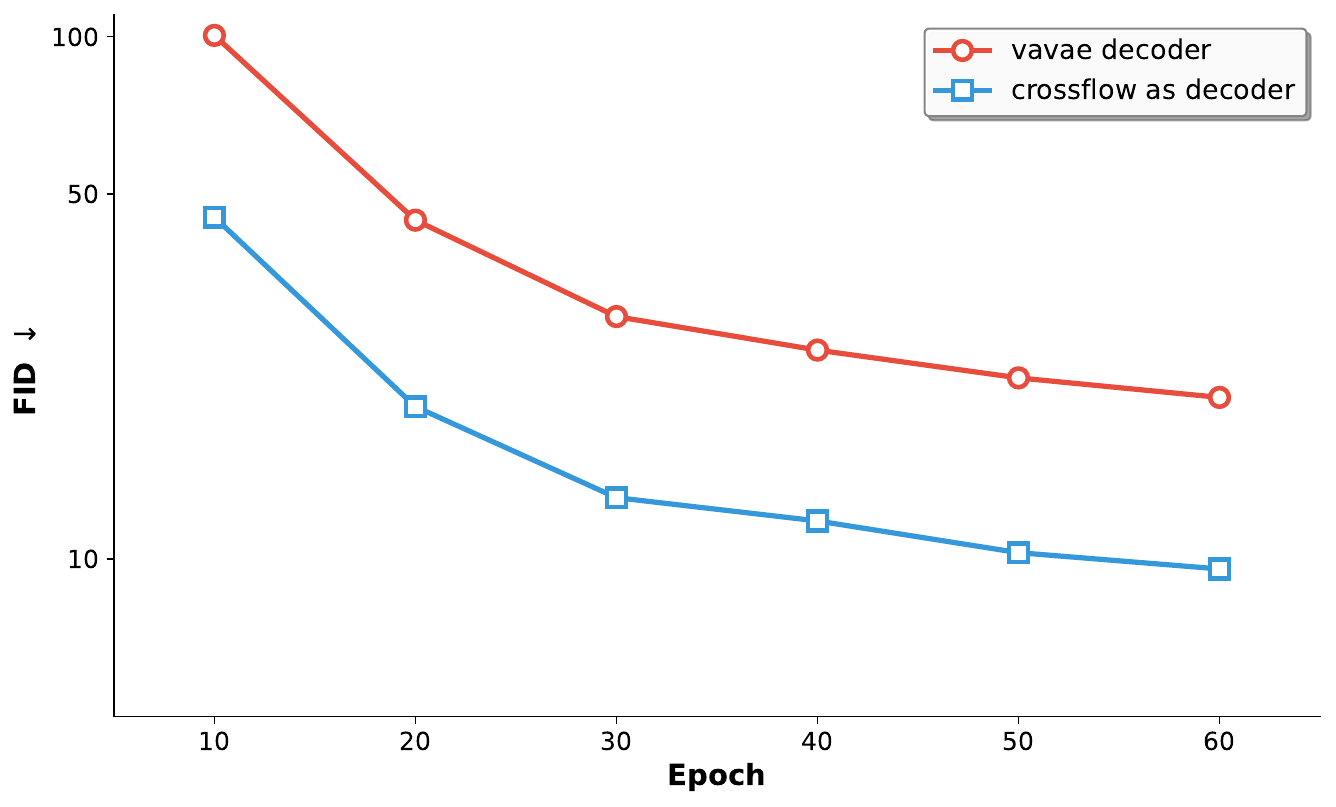}
    \captionof{figure}{\textbf{CrossFlow as a VAE decoder.} We evaluate CrossFlow as a VAE decoder for latent diffusion and report FID over training epochs for a LightningDiT-B/1 generator in the VA-VAE latent space. CrossFlow consistently improves generation quality over the VA-VAE decoder.}
    \label{fig:crossflow_vae}
\end{minipage}
\end{minipage}

\subsection{Performance as a VAE Decoder}

Finally, we evaluate CrossFlow as a decoder in a VAE-style pipeline. This setting tests whether the same latent-to-pixel model can serve not only as a generator, but also as a high-quality decoder for latents produced by a latent generator. We train LightningDiT-B/1 in the VA-VAE latent space~\citep{VAVAE}. Figure~\ref{fig:crossflow_vae} compares the same DiT generator with different decoders as training progresses. CrossFlow used as a VAE decoder improves FID throughout training, supporting the claim that the model can decode generator-produced latents more effectively than the default VA-VAE decoder in this setting.

\iffalse
\section{Limitations and Future Work}

The current study focuses on class-conditional ImageNet-1k at $256\times256$, so the results do not yet establish performance for text-to-image generation, substantially higher resolutions, or broader data domains. CrossFlow also depends on the quality of the fixed encoder: the encoder ablation shows that changing the latent representation materially affects FID. Finally, the strongest results use perceptual and adversarial losses, and the GAN term requires careful time-dependent weighting to avoid instability.

These limitations suggest several directions for future work. First, CrossFlow could be evaluated with stronger or jointly trained encoders to reduce the dependence on a fixed tokenizer. Second, text conditioning and higher-resolution generation would test whether the cross-space objective scales beyond class-conditional ImageNet. Third, alternatives to adversarial refinement may improve stability while retaining the benefit of direct pixel-space supervision.
\fi
\section{Conclusion}

We introduced \emph{CrossFlow}, a cross-space generative framework that maps latent noise directly to pixels. The central idea is to keep the efficiency of a latent input path while supervising the model with an image-valued one-step objective. Instantiated with a fixed encoder and a Transformer generator, CrossFlow unifies one-step generation and latent-to-pixel decoding in a single model. This design removes the separate pretrained decoder at inference, exposes generated images directly to perceptual and adversarial refinement losses, and improves decoding quality when input latents are produced by a latent generator. On ImageNet-1k at $256\times256$, CrossFlow-XL achieves 1.62 FID with a single function evaluation, and ablations show that the latent representation and pixel-space auxiliary losses are important for high-fidelity generation.

\newpage
{
    \small
    \bibliographystyle{IEEEtran}
    \bibliography{main}
}
\appendix

\newpage

\section{Broader Impact}

This work studies a general method for efficient image generation and does not introduce a domain-specific deployment. Potential benefits include lower inference cost and simpler latent-to-pixel generation pipelines, which may make high-quality synthesis more accessible for research and creative applications. At the same time, the method inherits standard risks of image generation systems, including the creation of misleading or harmful synthetic content and the reproduction of biases present in the training data. We therefore expect deployment to require the same safeguards as other generative image models, such as data curation, provenance or watermarking mechanisms, content filtering, and application-specific human oversight.

\section{Implementation Details}
\label{app:implementation_details}

This appendix provides implementation details that are useful for reproduction but not required to understand the main method. The main text summarizes the model design and the role of the auxiliary losses.

\subsection{Compute and Software}

All experiments are conducted on 8 Linux nodes, each with 8 NVIDIA H20 GPUs. Our implementation is based on PyTorch~\citep{pytorch}, Transformers~\citep{transformers_lib}, and timm~\citep{timm_lib}. We use torch-fidelity~\citep{torch_fid_lib} for FID computation. Training uses PyTorch Distributed Data Parallel with a per-GPU batch size of 32. We use the same fixed hyperparameter set across experiments rather than dense hyperparameter tuning.

\subsection{Architecture and Conditioning}

We use the VA-VAE encoder~\citep{VAVAE} as the fixed image encoder and a standard ViT backbone~\citep{ViT} as the CrossFlow generator. On ImageNet $256\times256$, the generator outputs $16\times16$ image patches, giving 256 image patch tokens. We additionally use 16 context tokens for class and time conditioning.

The theoretical formulation permits conditioning on both $t$ and $r$. In the implementation, we condition only on $r$. This choice follows the observation that a model can infer the noise level $t$ from the noisy latent input~\citep{DiffNoTime}, and it keeps the input format the same for reconstruction and one-step generation. Table~\ref{tab:ablation} reports the corresponding ablation: using both $(r,t)$ gives 2.73 FID, while the default $r$-only conditioning gives 2.54 FID.
\begin{center}
\captionof{table}{\textbf{Training setup for CrossFlow variants on ImageNet $256\times256$.} All variants use the same training schedule and differ mainly in depth, hidden dimension, and parameter count.}
\vspace{0.5em}
\small
\begin{tabular}{l c c c}
\toprule
\textbf{Hyperparameter} & \textbf{CrossFlow-B} & \textbf{CrossFlow-L}  & \textbf{CrossFlow-XL} \\
\midrule
Depth (layers) & 12 & 24 & 28\\
Hidden dimension & 768 & 1024 & 1152 \\
\# parameters & 95M & 315M & 461M\\
Patch size & 16 & 16  & 16 \\
\midrule
Epochs & 160 & 160 & 160 \\
Batch size & 2048 & 2048 & 2048 \\
Optimizer & Muon & Muon & Muon \\
Learning rate & 8e-4 & 8e-4 & 8e-4 \\
LR warmup & 5 epochs & 5 epochs & 5 epochs \\
EMA decay & \multicolumn{3}{c}{selected from 0.999/0.9995/0.9999} \\
Weight decay & 1e-6 & 1e-6 & 1e-6 \\
\bottomrule
\end{tabular}
\label{tab:hyperparams}
\end{center}

\subsection{Time Sampling}

For the time distribution, we start from a uniform distribution on $[0,1]$ and apply the shifted-time transformation used in RAE~\citep{RAE},
\begin{equation}
    t \mapsto \frac{\alpha t}{1 + (\alpha - 1)t},
    \qquad
    \alpha = \sqrt{\text{latent dimension}/4096}.
\end{equation}
This transformation adjusts the effective time distribution for different latent dimensionalities. With 50\% probability, we set $t=r$, in which case the CrossFlow loss reduces to the reconstruction endpoint. With the remaining 50\% probability, we sample two shifted times independently and assign them to $t$ and $r$ according to the time ordering used by the training objective.

\subsection{Auxiliary Pixel-Space Losses}

To improve perceptual quality, we combine the CrossFlow objective with perceptual and GAN losses following common VAE training practice~\citep{vqgan}. The perceptual loss is applied to the same scaled prediction and target used in the reconstruction-compatible objective:
\begin{equation}
\mathcal{L}_{\mathrm{perc}} =
\mathrm{PerceptualLoss}\!\left(
\hat{x} + \frac{\bigl(\beta(t)-\beta(r)\bigr)\beta(t)}{\dot{\beta}(t)\beta(r)}\,
\frac{d\hat{x}}{dt},
\Bigl[ \alpha(t) - \frac{\beta(t)}{\dot{\beta}(t)}\dot{\alpha}(t) \Bigr] x_0
\right).
\end{equation}

For adversarial training, we use a hinge GAN objective. The generator loss is
\begin{equation}
    \mathbb E_{t\sim T, z_t\sim p_t} -w(t)D(G(z_t, r)),
\end{equation}
and the discriminator loss is
\begin{equation}
    \mathbb E_{t\sim T, z_t\sim p_t, x\sim p}
    \max(1 - D(G(z_t, r)), 0) + \max(1 + D(x), 0).
\end{equation}
Here $w(t)$ controls the relative adversarial strength across time.

For the GAN discriminator architecture, we use the same architecture as dinoGAN in RAE~\citep{RAE}. For the perceptual loss, we map images to DINOv3-B features~\citep{dinov3} and compute a Huber loss between the feature tensors.

\subsection{GAN Weighting and Diagnostics}
\label{app:GANweight}

Because CrossFlow directly outputs pixels, adversarial training can be applied in the same space where image quality is evaluated. However, we observe that a uniform GAN weight can overemphasize low-signal regions of the trajectory. Figure~\ref{fig:gancollapse} illustrates the resulting collapse: when the vanilla GAN loss is used, low-time samples can ignore the class label and noise and collapse to similar images.

\begin{center}
    \includegraphics[width=0.45\linewidth]{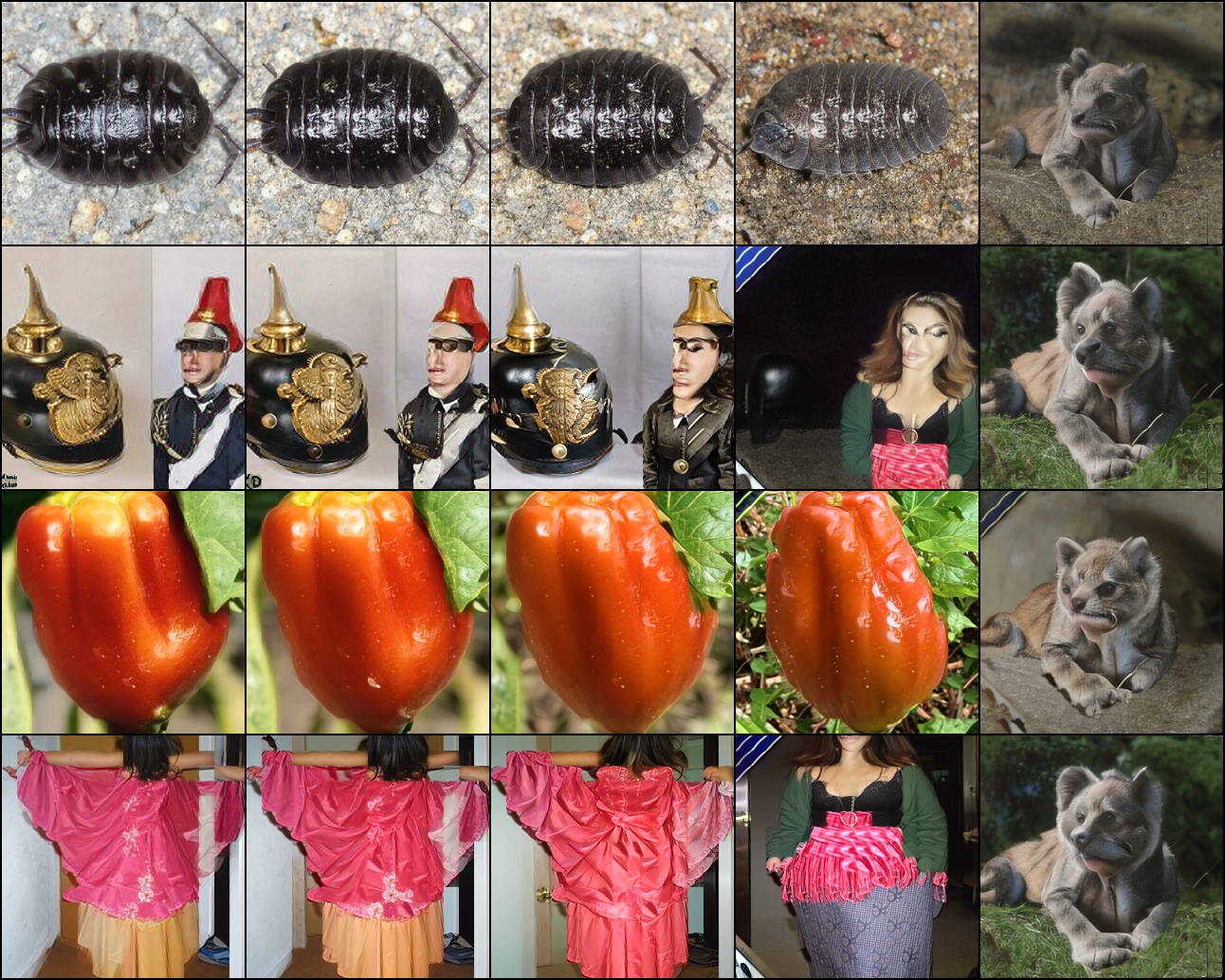}
    \captionof{figure}{Illustration of GAN collapse under a vanilla adversarial loss. Rows show different tuples of clean image, latent $z_1$, class label, and noise; columns show decreasing time values. The model remains reasonable in the high-signal region but can collapse at low time values.}
    \label{fig:gancollapse}
\end{center}

Figure~\ref{fig:gradnorm} compares the gradient norms of the GAN loss and the CrossFlow loss across time. The GAN gradient remains relatively stable, whereas the CrossFlow gradient varies with time, motivating a time-dependent adversarial weight.

\begin{center}
    \includegraphics[width=0.5\linewidth]{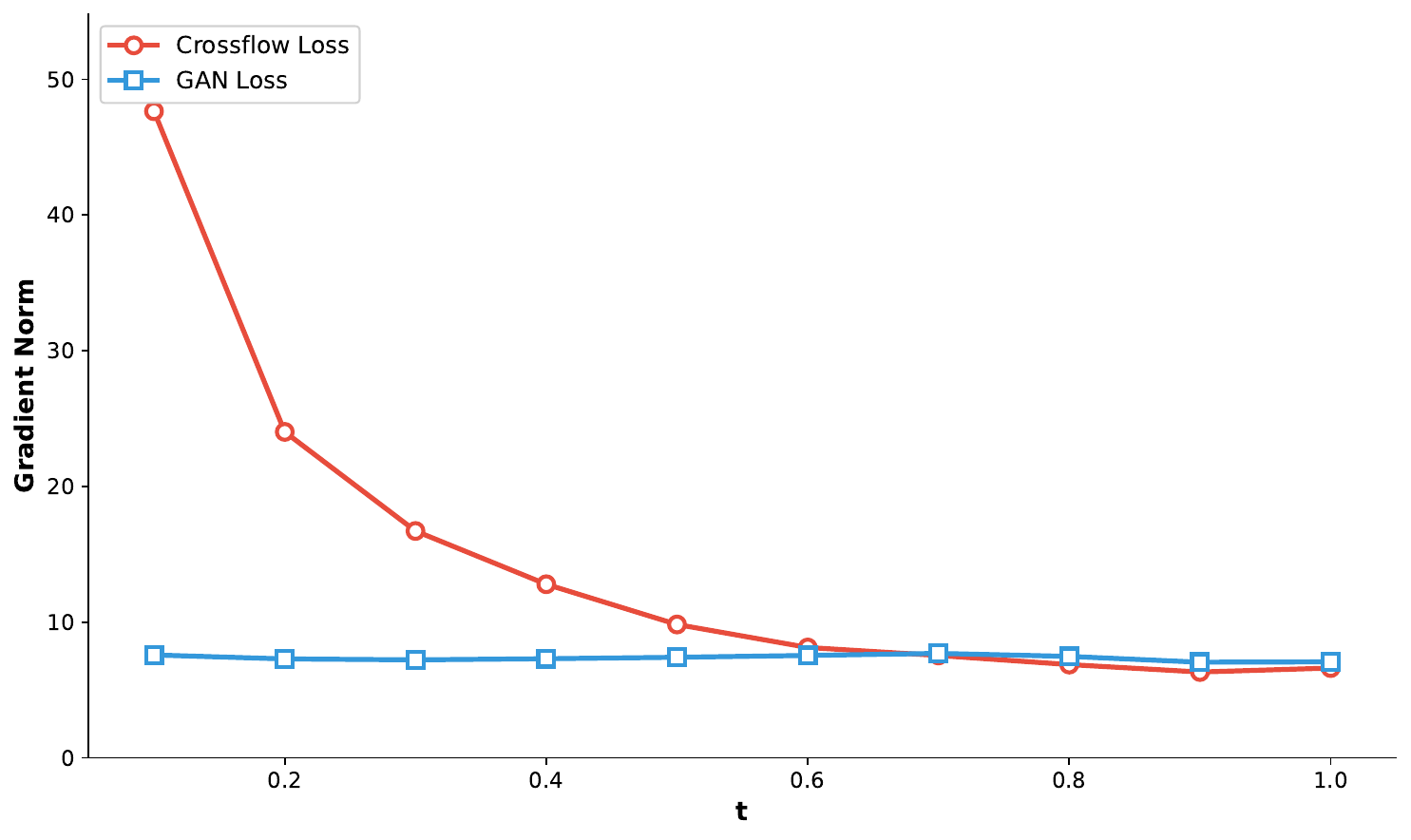}
    \captionof{figure}{Gradient norms of the GAN loss and CrossFlow loss across time, averaged over 204,800 training samples. The mismatch in relative gradient magnitude motivates weighting the adversarial loss by time.}
    \label{fig:gradnorm}
\end{center}

We therefore weight the generator loss by $w(t)$. We experiment with $w(t)=1-t$, $(1-t)^2$, and $(1-t)^4$, and use $(1-t)^2$ in the main experiments. Figure~\ref{fig:gan_weight} shows the resulting FID trajectories: too small a weight leads to weak adversarial refinement, while too large a weight can cause collapse.

\section{Additional Details on the CrossFlow Design Space}
\label{app:crossflow_design_space}

This appendix gives the details behind Sec.~\ref{sec:crossflow_design_space}. The purpose of the toy experiment is diagnostic: it checks which mathematical choices give a usable one-step map from noisy latents to images. It is not used as a main benchmark.

The CrossFlow objective Eq.~\eqref{eq:crossflow_loss_family} contains several time-dependent coefficients. The noise schedule $\alpha,\beta$ defines the latent probability path, while $\phi,\lambda,\gamma$ determine the one-step statistic and its induced training residual. The natural design order is:
\[
\text{choose } \phi(r)
\quad\Longrightarrow\quad
\text{choose } \lambda(t,r)
\quad\Longrightarrow\quad
\text{derive } \gamma(t,r).
\]
Indeed, the velocity-cancellation condition Eq.~\eqref{eq:cancel_v} gives
\begin{equation}
\gamma(t,r)\lambda(t,r)
=
1-\frac{\phi(r)}{\beta(t)}.
\label{eq:app_cancel_v}
\end{equation}
Therefore, once $\phi$ and $\lambda$ are chosen, the original coefficient $\gamma$ is determined by
\begin{equation}
\gamma(t,r)
=
\frac{1-\phi(r)/\beta(t)}{\lambda(t,r)}.
\label{eq:app_gamma_from_phi_lambda}
\end{equation}

When $\phi(r)\neq0$, we can divide the residual by $\phi(r)$. We denote the divided coefficient by
\begin{equation}
\tilde{\gamma}(t,r)
:=
\frac{\gamma(t,r)}{\phi(r)}
=
\frac{1}{\lambda(t,r)}
\left(
\frac{1}{\phi(r)}
-
\frac{1}{\beta(t)}
\right).
\label{eq:app_tilde_gamma_from_phi_lambda}
\end{equation}
Thus, after Observation~1, the practical design space is most clearly discussed in terms of $\tilde{\gamma}$, while the original coefficient is always recovered as
\begin{equation}
\gamma(t,r)=\phi(r)\tilde{\gamma}(t,r).
\label{eq:app_gamma_recover}
\end{equation}

We use a Swiss-roll toy problem as a diagnostic. All variants use the same MLP, the same fixed encoder, the same detached directional derivative, the same paired-data conditional velocity, and 800 training steps. Lower sliced Wasserstein distance (SWD) is better. The toy implementation uses the reverse time convention, where $t=0$ is noise and $t=1$ is the clean latent; for readability, the formulas below are written in the paper convention.

\begin{table}[t]
\centering
\caption{Swiss-roll diagnostic for the main design choices. In O2 and O3, the variants are written in terms of the divided coefficient $\tilde{\gamma}=\gamma/\phi$. The original coefficient is $\gamma=\phi\tilde{\gamma}$. The cosine rows use $\alpha^2+\beta^2=1$; all other rows use the linear schedule. Lower SWD is better.}
\label{tab:design_questions_main}
\small
\resizebox{\linewidth}{!}{%
\begin{tabular}{l l c}
\hline
Observation & Variant & SWD $\downarrow$ \\
\hline
O1 & $\phi(r)=\beta(r)$, $\lambda=1$, hence $\gamma(t,r)=1-\frac{\beta(r)}{\beta(t)}$ and $\tilde{\gamma}(t,r)=\frac{1}{\beta(r)}-\frac{1}{\beta(t)}$ & $0.0087$ \\
O1 & $\phi=0$, no boundary condition & $0.4483$ \\
O1 & $\phi=0$, clean boundary only & $0.3938$ \\
\hline
O2 & $\tilde{\gamma}(t,r)=\frac{1}{\beta(t)}$, never zero & $0.2775$ \\
O2 & $\tilde{\gamma}(t,r)=\frac{1}{\beta(r)}-\frac{1}{\beta(t)}$, diagonal zero $\tilde{\gamma}(t,t)=0$ & $0.0087$ \\
O2 & $\tilde{\gamma}_\delta(t,r)=\frac{1}{\beta(r+0.1)}-\frac{1}{\beta(t)}$, shifted zero $\tilde{\gamma}_\delta(t,t-0.1)=0$ & $0.0169$ \\
O2 & $\tilde{\gamma}_\delta(t,r)=\frac{1}{\beta(r+0.2)}-\frac{1}{\beta(t)}$, shifted zero $\tilde{\gamma}_\delta(t,t-0.2)=0$ & $0.0455$ \\
\hline
O3 & linear schedule, inverse-$\beta$ $\tilde{\gamma}$, matched & $0.0087$ \\
O3 & linear schedule, cosine-matched $\tilde{\gamma}(t,r)=\cot\frac{\pi r}{2}-\cot\frac{\pi t}{2}$, not matched & $0.3829$ \\
O3 & cosine schedule, inverse-$\beta$ $\tilde{\gamma}$, not matched & $0.0125$ \\
O3 & cosine schedule, $\tilde{\gamma}(t,r)=\cot\frac{\pi r}{2}-\cot\frac{\pi t}{2}$, matched & $0.0094$ \\
\hline
\end{tabular}%
}
\end{table}

\paragraph{Observation 1: choose $\phi(r)\neq0$ so the loss contains the data term.}
The first question is whether we can set $\phi(r)$ to the constant zero function. If $\phi(r)=0$, Eq.~\eqref{eq:app_cancel_v} gives
\begin{equation}
\gamma(t,r)\lambda(t,r)=1.
\end{equation}
In this case, the image-valued data term disappears from the residual:
\begin{equation}
\mathcal L_{\phi=0}
=
\mathbb E
\left\|
\partial_t\gamma(t,r)F_\theta(z_t,t,r)
+
\gamma(t,r)\frac{\mathrm d F_\theta}{\mathrm dt}
\right\|_2^2 .
\label{eq:cons_branch}
\end{equation}
For example, with $\gamma(t,r)=1$, the loss only asks the prediction to remain constant along the latent trajectory. It does not specify which image should be predicted. Therefore, the $\phi=0$ branch needs an additional boundary loss, similar to consistency models. The toy experiment confirms this behavior: with no boundary, the model is essentially unanchored; with only a clean boundary, generation from noise is still poor. In contrast, the branch $\phi(r)\neq0$ performs much better because the residual itself contains the image-valued data term.

When $\phi(r)=0$, Eq.~\eqref{eq:app_cancel_v} gives $\gamma(t,r)\lambda(t,r)=1$, and the one-space statistic becomes
\begin{equation}
Z(z_t, t, r; \lambda, \gamma)
=
\lambda(t, r)
\left(
z_t
+
\int_{t}^{r} v(z_\tau,\tau)\,\mathrm d\tau
\right),
\label{eq:consistency_statistic}
\end{equation}
which is a scaled consistency-model statistic. This branch asks the prediction to remain consistent along the probability-flow trajectory. Its cross-space loss is
\begin{equation}
\mathbb E_{x_0,\epsilon,t,r}
\left\|
\partial_t\gamma(t,r)F_\theta(z_t,t,r)
+
\gamma(t,r)\frac{\mathrm dF_\theta}{\mathrm dt}
\right\|_2^2 .
\label{eq:consistency_like_loss}
\end{equation}
For $\gamma(t,r)=1$, this reduces to the standard consistency-style constraint
\begin{equation}
\mathbb E_{x_0,\epsilon,t,r}
\left\|
\frac{\mathrm dF_\theta}{\mathrm dt}
\right\|_2^2 .
\label{eq:standard_consistency_loss}
\end{equation}
The constraint keeps the image prediction unchanged along the trajectory, but it does not identify which image should be predicted. A trivial constant output can satisfy it. This branch therefore needs an additional reconstruction boundary,
\begin{equation}
\mathbb E_{x_0\sim p_{\mathrm{data}}}
\left\|
F_\theta(\mathcal E(x_0),0,0)-x_0
\right\|_2^2 .
\label{eq:reconstruction_boundary}
\end{equation}
This makes the branch a minimal cross-space consistency baseline: the flow loss enforces invariance, while the boundary loss defines the invariant prediction.

There is also a boundary subtlety. Two-point consistency models in one space use the noisy-state boundary $Z(z_t,t,t)=\gamma(t,t)z_t$. In cross-space generation, this boundary has no direct analogue because $z_t$ is latent-valued but the desired output is image-valued. We keep this branch as a meaningful baseline, but not as the preferred CrossFlow objective.

We therefore use the branch $\phi(r)\neq0$. Since $\phi$ depends only on $r$, dividing the residual by $\phi(r)$ does not introduce additional $t$-derivative terms. Using $\tilde{\gamma}=\gamma/\phi$, the divided residual can be written as
\begin{equation}
\mathcal L_{\phi\neq0}
=
\mathbb E
\Bigg\|
\partial_t\tilde{\gamma}(t,r)F_\theta(z_t,t,r)
+
\tilde{\gamma}(t,r)\frac{\mathrm dF_\theta}{\mathrm dt}
-
\frac{\dot\beta(t)}{\beta(t)^2}
\left(
\alpha(t)-\dot\alpha(t)\frac{\beta(t)}{\dot\beta(t)}
\right)x_0
\Bigg\|_2^2 .
\label{eq:neq_loss}
\end{equation}
Multiplying the whole residual by $-1$ gives an equivalent squared loss; Eq.~\eqref{eq:neq_loss} uses the sign convention induced by $\tilde{\gamma}=\gamma/\phi$. After this simplification, the remaining design questions are how to choose $\lambda$ so that the resulting $\tilde{\gamma}$ has useful zeros, and how to match $\tilde{\gamma}$ to the noise schedule.

\paragraph{Observation 2: choose $\lambda$ so the resulting $\tilde{\gamma}$ has diagonal reconstruction anchors.}
The second question is how to choose $\lambda(t,r)$ after fixing a nonzero $\phi(r)$. From Eq.~\eqref{eq:app_tilde_gamma_from_phi_lambda},
\begin{equation}
\tilde{\gamma}(t,r)
=
\frac{1}{\lambda(t,r)}
\left(
\frac{1}{\phi(r)}
-
\frac{1}{\beta(t)}
\right).
\label{eq:app_tilde_gamma_after_lambda}
\end{equation}
Since $\tilde{\gamma}(t,r)$ multiplies $\mathrm dF_\theta/\mathrm dt$ in Eq.~\eqref{eq:neq_loss}, the derivative term vanishes wherever $\tilde{\gamma}(t,r)=0$. At such points, Eq.~\eqref{eq:neq_loss} reduces to
\begin{equation}
\mathbb E
\Bigg\|
\partial_t\tilde{\gamma}(t,r)F_\theta(z_t,t,r)
-
\frac{\dot\beta(t)}{\beta(t)^2}
\left(
\alpha(t)-\dot\alpha(t)\frac{\beta(t)}{\dot\beta(t)}
\right)x_0
\Bigg\|_2^2 .
\label{eq:recon_loss}
\end{equation}
Thus, a zero of $\tilde{\gamma}$ turns the CrossFlow residual into a reconstruction-type loss.

The most natural choice is to place this zero on the diagonal $r=t$. If $\lambda(t,r)$ is finite and nonzero, then Eq.~\eqref{eq:app_tilde_gamma_after_lambda} gives
\begin{equation}
\tilde{\gamma}(t,t)=0
\quad\Longleftrightarrow\quad
\phi(t)=\beta(t).
\label{eq:app_diagonal_phi_condition}
\end{equation}
Therefore, the simplest diagonal-anchor construction is
\begin{equation}
\phi(r)=\beta(r),
\qquad
\lambda(t,r)=1.
\label{eq:app_simple_phi_lambda}
\end{equation}
Substituting Eq.~\eqref{eq:app_simple_phi_lambda} into Eq.~\eqref{eq:app_gamma_from_phi_lambda} gives the original statistic coefficient
\begin{equation}
\gamma(t,r)
=
1-\frac{\beta(r)}{\beta(t)},
\label{eq:app_original_gamma_main}
\end{equation}
and the divided coefficient
\begin{equation}
\tilde{\gamma}(t,r)
=
\frac{\gamma(t,r)}{\phi(r)}
=
\frac{1}{\beta(r)}
-
\frac{1}{\beta(t)} .
\label{eq:gamma_inv}
\end{equation}

The toy results show why the diagonal zero matters. A simple nonzero choice, such as $\tilde{\gamma}(t,r)=1/\beta(t)$, performs poorly because the derivative term is always present and the objective never becomes a pure reconstruction anchor. Shifted-zero variants also create reconstruction anchors, but the anchors are moved away from the natural pair $(t,t)$. Their performance degrades as the shift grows. We therefore use the diagonal condition
\begin{equation}
\tilde{\gamma}(t,t)=0,
\label{eq:diagonal_zero_condition}
\end{equation}
which gives a direct reconstruction anchor at every noise level and keeps the objective easy to interpret.

\paragraph{Observation 3: choose $\phi$ and $\lambda$ jointly with the noise schedule so $\tilde{\gamma}$ has the correct scale.}
Eq.~\eqref{eq:recon_loss} shows that, when $\tilde{\gamma}(t,t)=0$, the diagonal residual becomes a reconstruction loss. However, it reconstructs $x_0$ itself only if the coefficient of $F_\theta$ matches the coefficient of $x_0$. This requires
\begin{equation}
\partial_t\tilde{\gamma}(t,t)
=
\frac{\dot\beta(t)}{\beta(t)^2}
\left(
\alpha(t)-\dot\alpha(t)\frac{\beta(t)}{\dot\beta(t)}
\right).
\label{eq:diagonal_scale_condition}
\end{equation}
Otherwise, the model is asked to fit a scaled version of $x_0$, which makes the supervision less direct.

A convenient construction is to choose a scalar function $g$ and set
\begin{equation}
\tilde{\gamma}(t,r)
=
g(r)-g(t),
\label{eq:app_tilde_gamma_g_difference}
\end{equation}
where
\begin{equation}
g'(t)
=
-\frac{\dot\beta(t)}{\beta(t)^2}
\left(
\alpha(t)-\dot\alpha(t)\frac{\beta(t)}{\dot\beta(t)}
\right).
\label{eq:schedule_matched_gamma}
\end{equation}
Then $\tilde{\gamma}(t,t)=0$, and
\begin{equation}
\partial_t\tilde{\gamma}(t,t)
=
-g'(t)
=
\frac{\dot\beta(t)}{\beta(t)^2}
\left(
\alpha(t)-\dot\alpha(t)\frac{\beta(t)}{\dot\beta(t)}
\right),
\end{equation}
so Eq.~\eqref{eq:diagonal_scale_condition} is satisfied.

This desired $\tilde{\gamma}$ can be implemented in the original design order. First choose
\begin{equation}
\phi(r)=\frac{1}{g(r)} .
\label{eq:app_phi_from_g}
\end{equation}
Then choose
\begin{equation}
\lambda(t,r)
=
\frac{
g(r)-\frac{1}{\beta(t)}
}{
g(r)-g(t)
}.
\label{eq:app_lambda_from_g}
\end{equation}
Substituting Eqs.~\eqref{eq:app_phi_from_g} and~\eqref{eq:app_lambda_from_g} into Eq.~\eqref{eq:app_tilde_gamma_from_phi_lambda} gives
\begin{equation}
\begin{aligned}
\tilde{\gamma}(t,r)
&=
\frac{1}{\lambda(t,r)}
\left(
\frac{1}{\phi(r)}
-
\frac{1}{\beta(t)}
\right) \\
&=
\frac{g(r)-g(t)}
{g(r)-\frac{1}{\beta(t)}}
\left(
g(r)
-
\frac{1}{\beta(t)}
\right) \\
&=
g(r)-g(t).
\end{aligned}
\label{eq:app_tilde_gamma_recovered}
\end{equation}
The original coefficient is then
\begin{equation}
\gamma(t,r)
=
\phi(r)\tilde{\gamma}(t,r)
=
1-\frac{g(t)}{g(r)} .
\label{eq:app_original_gamma_from_g}
\end{equation}

For the linear schedule used in our main experiments, Eq.~\eqref{eq:schedule_matched_gamma} gives
\begin{equation}
g(t)=\frac{1}{\beta(t)} .
\end{equation}
Therefore,
\begin{equation}
\phi(r)=\beta(r),
\qquad
\lambda(t,r)=1,
\qquad
\gamma(t,r)=1-\frac{\beta(r)}{\beta(t)},
\qquad
\tilde{\gamma}(t,r)=\frac{1}{\beta(r)}-\frac{1}{\beta(t)} .
\end{equation}
This recovers the simple CrossFlow construction. In practice, we clip the endpoint singularity for numerical stability.

For schedules that do not satisfy $\alpha+\beta=1$, inverse-$\beta$ is no longer automatically schedule-matched. For example, for the cosine schedule
\begin{equation}
\alpha(t)=\cos\frac{\pi t}{2},
\qquad
\beta(t)=\sin\frac{\pi t}{2},
\qquad
\alpha(t)^2+\beta(t)^2=1,
\end{equation}
Eq.~\eqref{eq:schedule_matched_gamma} gives
\begin{equation}
g(t)=\cot\frac{\pi t}{2}.
\end{equation}
A schedule-matched construction is therefore
\begin{equation}
\phi(r)=\tan\frac{\pi r}{2},
\qquad
\lambda(t,r)
=
\frac{
\cot\frac{\pi r}{2}
-
\frac{1}{\sin(\pi t/2)}
}{
\cot\frac{\pi r}{2}
-
\cot\frac{\pi t}{2}
},
\end{equation}
which yields
\begin{equation}
\tilde{\gamma}_{\cos}(t,r)
=
\cot\frac{\pi r}{2}
-
\cot\frac{\pi t}{2}.
\label{eq:gamma_cos}
\end{equation}
In the Swiss-roll diagnostic, this schedule-matched cosine choice improves over raw inverse-$\beta$ and gives the best cosine-schedule result.

Overall, the design rule is simple. First choose $\phi(r)\neq0$ so that the residual contains an image-valued target. Then choose $\lambda(t,r)$ to satisfy the desired endpoint behavior and to place reconstruction anchors at useful locations. Finally, $\gamma$ is determined by Eq.~\eqref{eq:app_gamma_from_phi_lambda}, and the divided residual is governed by $\tilde{\gamma}=\gamma/\phi$. For the linear schedule, this logic gives $\phi(r)=\beta(r)$, $\lambda(t,r)=1$, $\gamma(t,r)=1-\beta(r)/\beta(t)$, and $\tilde{\gamma}(t,r)=1/\beta(r)-1/\beta(t)$, with mild endpoint clipping for numerical stability.
\end{document}